%% file: camera_ready.tex
\documentclass{article}

\usepackage{microtype}
\usepackage{graphicx}
\usepackage{subcaption}
\usepackage{booktabs} 

\usepackage{hyperref}

\usepackage[accepted]{icml2026}

\usepackage{amsmath}
\usepackage{amssymb}
\usepackage{mathtools}
\usepackage{amsthm}

\usepackage{caption}        
\usepackage{enumitem}       

\usepackage[capitalize,noabbrev]{cleveref}

\icmltitlerunning{A Fair Evaluation of Graph Foundation Models for Node Property Prediction}

\begin{document}

\twocolumn[
  \icmltitle{A Fair Evaluation of Graph Foundation Models for Node Property Prediction}



  \icmlsetsymbol{equal}{*}

  \begin{icmlauthorlist}

    \icmlauthor{Oleg Platonov}{hse,yr}
    \icmlauthor{Gleb Bazhenov}{hse,yr}
    \icmlauthor{Dmitry Eremeev}{hse,yr}
    \icmlauthor{Liudmila Prokhorenkova}{yr}
    
  \end{icmlauthorlist}

  \icmlaffiliation{hse}{HSE University}
  \icmlaffiliation{yr}{Yandex Research}

  \icmlcorrespondingauthor{Oleg Platonov}{olegplatonov@yandex-team.ru}

  \icmlkeywords{Graph Foundation Model, GFM, Graph Neural Network, GNN, Node Property Prediction, Evaluation, Benchmarking}

  \vskip 0.3in
]



\printAffiliationsAndNotice{}  

\begin{abstract}
  Due to the wide use of graph-structured data in different fields of industry and science, the development of Graph Foundation Models (GFMs) has recently attracted a lot of attention. While many different types of models are called GFMs, particular interest has been paid to GFMs designed for node property prediction tasks, which is one of the most popular settings in Graph ML with lots of real-world applications from fraud detection in financial and social networks to recommendation systems for e-commerce and user-generated content platforms. While a number of GFMs for this task have been recently proposed, the field has not converged to a unified evaluation setting, and different works evaluate their models in widely different ways, preventing reliable comparison of GFMs with each other and with other types of models. In this work, we conduct a fair and rigorous reevaluation of 9 recent GFMs for node property prediction, comparing them to strong Graph Neural Network (GNN) baselines. We find that, among these GFMs, only the most recent ones based on the Prior-data Fitted Networks paradigm outperform well-tuned GNNs in predictive performance, although at a higher inference cost.
\end{abstract}

\section{Introduction}

Graphs are a natural way to represent many types of data, and thus Graph Machine Learning (i.e., applying machine learning to graph-structured data) has recently gained a lot of attention. In the past decade, message-passing Graph Neural Networks (GNNs) \citep{duvenaud2015convolutional, kipf2017semi, gilmer2017neural, hamilton2017inductive} have emerged as a particularly effective approach to many tasks in Graph Machine Learning and become the default model type in the field. One of the most common tasks in Graph Machine Learning, and one in which GNNs have been particularly successful, is node property prediction, i.e., predicting unknown labels of nodes in a large graph from node features, graph structure, and known node labels. This task attracts a lot of attention as it has many impactful real-world applications: large-scale industrial use cases of node property prediction include applying GNNs to social, financial, co-purchasing, content-similarity, and transportation networks to solve tasks like fraud detection, churn prediction, content or product search and recommendation, click-through rate prediction, estimated time of arrival prediction, and many others \citep{ying2018graph, wang2019semi, chen2021structured, derrow2021eta, fey2023relational, de2024personalized, borisyuk2024lignn, zhao2025gigl, GraphLand}.

GNNs are typically trained from scratch for each specific task. However, recently some machine learning fields have achieved success in developing Foundation Models~--- large multi-purpose pretrained models capable of solving many different tasks. Thus, while classic GNNs currently remain the dominant approach to Graph Machine Learning, there have recently been many attempts at developing Graph Foundation Models (GFMs). While different types of GFMs can support different tasks and data domains, due to high interest in node property prediction applications, a number of GFMs have been designed specifically for node property prediction across various datasets \citep{OFA, li2024zerog, OpenGraph, AnyGraph, GCOPE, SAMGPT, MDGFM, UniGraph, TSGNN, GraphFM, G2TFM, eremeev2026graphpfn, choi2025can, NodePFN, TAG}. However, despite many GFMs for node property prediction being developed, the community has not currently adopted a unified evaluation setting for these models and different works perform evaluation in widely different ways. This prevents reliable comparison of GFMs with each other and with other model types. Further, many evaluation setups in the literature have such issues as selecting narrow and unrepresentative sets of datasets and using weak baselines.

To obtain a reliable assessment of the current state of the field, we conduct a large empirical study that reevaluates 9 recent GFMs for node property prediction in a more reliable and realistic setting. We find that the predictive performance of different models differs drastically, and most of them cannot outperform properly tuned GNNs. Only several recent GFMs based on the Prior-data Fitted Networks (PFNs) paradigm \citep{muller2022transformers} (this paradigm first gained success in developing Foundation Models for tabular data \citep{TabPFN, TabPFNv2}) achieve strong results and are able to outperform GNNs, although these models also come \hbox{with significantly higher computational costs of inference}.

\vspace{-5pt}

\section{Background}

\vspace{-3pt}

The term Foundation Model refers to a machine learning model that can be applied to a wide range of tasks with no or minimal adaptation \citep{bommasani2021opportunities}. Such models are commonly pretrained on large amounts of data and then applied to downstream tasks with relatively light fine-tuning or even without any fine-tuning at all (e.g., in the in-context learning regime). Foundation Models have achieved great success in the fields of Natural Language Processing, Computer Vision, and Audio Processing, transforming these fields and becoming default approaches in them. This has led to many attempts to design Foundation Models for other types of data, including graph-structured data. However, we note that graph-structured data differs significantly from text, image, video, or audio data in that it can arguably be better viewed not as a single data domain, but rather a general way to represent data from different domains. Indeed, there is not much in common between graphs representing social networks, road networks, databases, ontologies, molecules, connectomes, protein-protein interaction networks, and gene regulatory networks (and this is just a small set of examples of data types commonly represented as graphs). The practical tasks that these graphs are used to solve are also very different and include predicting properties of graph components (nodes, edges, or more complex graph substructures) or of entire graphs, generating graphs or their components, and even predicting interactions between multiple graphs (e.g., molecules). Further, even a single task type can have very different meanings and represent entirely different real-world problems in different graphs (e.g., predicting whether a user is a fraudster in a social network and whether an accident is likely on a road segment in a transportation network are both node property prediction tasks) and even in the same graph (e.g., predicting whether a user is a fraudster in a social network and estimating the likelihood of a user leaving the platform, i.e., churn prediction, in the same social network). This has led to the natural result that, while many models currently proposed in the literature are called GFMs, they all target different types of graphs and/or different tasks. For example, considering data domain, there are GFMs meant only for molecular graphs \citep{beaini2024towards}, only for knowledge graphs \citep{galkin2024towards}, or only for road graphs \citep{wang2023building}. Further, considering task type, there are GFMs meant only for node property prediction \citep{G2TFM, eremeev2026graphpfn}, only for graph property prediction \citep{frasca2024towards}, or only for link prediction \citep{dong2024universal}. We thus argue that the term GFM is not well-defined in itself and encourage researchers to explicitly clarify the scope of applicability of their model when calling it a GFM.

\vspace{-2pt}

In this work, we focus on GFMs for node property prediction. Specifically, in node property prediction, given a graph and labels of some of the nodes, the model should predict labels for the rest of the nodes. The labels can be either categorical (node classification) or continuous (node regression). This task appears in many real-world applications, and many GFMs for this task have been proposed. The key challenge for GFMs dealing with node property prediction is that graphs can come from vastly different domains (e.g., social networks, co-purchasing networks, road networks). This implies that the graphs can have different node feature spaces and node target spaces, i.e., have features and targets with different dimensions, distributions, meanings, and relations to each other. Thus, a successful GFM for node property prediction must be able to effectively adapt to graphs with different node feature and target spaces. The GFMs being proposed in the literature differ significantly in their mechanisms for this adaptation~--- we discuss some of these mechanisms in Appendix~\ref{app:gfms}. However, we note that the field of tabular deep learning faces the same challenge of having different feature and target spaces for different datasets, and in this field several successful Foundation Models \citep{TabPFN, TabPFNv2, qu2025tabicl, qu2026tabiclv2, LimiX} have recently been created following the Prior-data Fitted Networks (PFNs) paradigm \citep{muller2022transformers}. This paradigm trains Transformer-based \citep{vaswani2017attention} models on a large number of synthetic datasets to learn to process the entire dataset and adapt to the new feature and target spaces in a single forward pass and thus make predictions in the in-context learning regime (we describe the PFNs paradigm in more detail in Appendix~\ref{app:pfns}). Several recent works \citep{G2TFM, eremeev2026graphpfn, choi2025can, NodePFN, TAG} have adapted the PFNs paradigm to GFMs for node property prediction.

\vspace{-5pt}

\section{Experiments}
\label{sec:experiments}

\vspace{-1pt}

\subsection{Datasets}
\label{sec:datasets}

\vspace{-1pt}

Recently, there has been a lot of criticism of datasets commonly used in Graph ML due to their questionable practical relevance, narrow domain coverage, bugs in the data collection process, and other issues \citep{platonov2023critical, li2023graphcleaner, bechler2025position, GraphLand} (we provide more details on these issues in Appendix~\ref{app:datasets}). Thus, for our experiments, we use $10$ datasets from the recently introduced GraphLand benchmark \citep{GraphLand} that represents real-world applications of node property prediction and was designed to address the issues of prior datasets. We use the \texttt{RL} data splits. To aggregate results across multiple datasets, we compute average model rank and average normalized score (a normalized score is calculated by linearly transforming metrics obtained on a specific dataset so that the worst model achieves a value of $0$ and \hbox{the best model achieves a value of $1$)}.

\vspace{-4pt}

\subsection{Models}

\vspace{-4pt}

\paragraph{Graph Neural Networks}
As noted in the broader ML literature \citep{lipton2018troubling}, baseline tuning often receives less attention than tuning of newly proposed models, and Graph ML research is not immune to this tendency. It is often reported that newer methods outperform classic GNNs, however, a number of works over the years have performed more fair reevaluations of classic GNNs \citep{shchur2018pitfalls, errica2020fair, platonov2023critical, tonshoff2023did, CGASB, CGASB2, GraphLand} in different settings and found that their performance can be substantially improved by performing proper hyperparameter search and adding standard deep learning building blocks such as skip-connections \citep{he2016deep} and normalizations \citep{ioffe2015batch, ba2016layer}. In particular, \citet{platonov2023critical} proposed stronger GNN variants with architectural improvements, and \citet{CGASB} recently provided a relatively simple but strong evaluation protocol for node classification with GNNs including both hyperparameter search and architectural improvements. In our work, to provide a fair model comparison, we use two versions of strong GNNs with improved backbones~--- those from \citet{platonov2023critical} and those from \citet{CGASB}. We conduct extensive hyperparameter search for them: $100$ trials with the TPE algorithm \citep{bergstra2011algorithms, watanabe2023tree}. More specifically, we evaluate the improved versions of 4 GNN models: GCN \citep{kipf2017semi}, GraphSAGE \citep{hamilton2017inductive}, GAT \citep{velivckovic2018graph}, and Local Graph Transformer (LGT) \citep{shi2021masked, platonov2023critical, platonov2026cluster}. We provide more details on these models in Appendix~\ref{app:gnns}.

\vspace{-8pt}

\paragraph{Graph Foundation Models}
We evaluate the following 9 GFMs: OpenGraph \citep{OpenGraph}, AnyGraph \citep{AnyGraph}, GCOPE \citep{GCOPE}, SAMGPT \citep{SAMGPT}, MDGFM \citep{MDGFM}, TS-GNN \citep{TSGNN}, G2T-FM \citep{G2TFM}, TAG \citep{TAG}, GraphPFN \citep{eremeev2026graphpfn}. We use their open-source official implementations for all models. G2T-FM, TAG, and GraphPFN are based on the PFNs paradigm. GraphPFN uses a custom graph-native architecture, while G2T-FM and TAG directly use unmodified Tabular Foundation Models as backbones. For G2T-FM, we evaluate the LimiX-16M \citep{LimiX} backbone, which the authors of G2T-FM recommend as the best performing one. For TAG, we evaluate both the TabPFNv2 \citep{TabPFNv2} and LimiX-16M \citep{LimiX} backbones, which achieve the best results in the original TAG paper. GCOPE, SAMGPT, and MDGFM only support the fine-tuning (FT) regime. OpenGraph, AnyGraph, TS-GNN, and all PFN-based GFMs support in-context learning (ICL), and the official implementations of G2T-FM and GraphPFN also support fine-tuning. We evaluate each model using all the supported regimes.
Note that some of the considered models do not support node regression~--- we evaluate these models only on node classification datasets.

\vspace{-4pt}

\subsection{Results: Predictive Performance}

\begin{table*}[t]
\centering
\captionsetup{font=footnotesize}
\caption{Experimental results. Accuracy, AP, and $R^2$ are reported for mult. class., bin. class., and regression datasets, respectively. Average rank (lower is better) and average normalized score (higher is better) are computed either only across classification datasets (\textit{rank (cls)} and \textit{score (cls)}) or across all datasets (\textit{rank (all)} and \textit{score (all)}). The \textit{Crit.} and \textit{Class.} prefixes refer to improved GNNs from \citet{platonov2023critical} and \citet{CGASB}, respectively \hbox{(by the first words of the paper titles). The best results are highlighted: \textcolor{red}{first}, \textcolor{orange}{second}, \textcolor{violet}{third}}.}
\vspace{-6pt}
\input{inputs/tables/results}
\label{tab:experimental-results}
\vspace{-4pt}
\end{table*}

\begin{table*}[h]
\centering
\captionsetup{font=footnotesize}
\caption{Time required for: hyperparameter tuning (Tun), a single training run with the best hyperparameters (Tr), a single inference run (including inference-time ensembling for PFN-based models) with the best hyperparameters (Inf) using NVIDIA
Tesla A100 80GB GPU. Note that for GNNs, the models with the best hyperparameters can be widely different in size across GNN types and datasets.}
\vspace{-6pt}
\input{inputs/tables/total_time}
\label{tab:time}
\vspace{-16pt}
\end{table*}

\vspace{-4pt}

We describe the details of our experimental setup in Appendix~\ref{app:experimental-setup}. Our experimental results are provided in Table~\ref{tab:experimental-results}. Based on them, we make the following observations:

\vspace{-10pt}

\begin{itemize}[leftmargin=6pt]
    \item The results clearly separate GFMs into two groups, which correspond to PFN-based models and non-PFN-based models. Non-PFN-based GFMs always underperform well-tuned classic GNNs on the considered datasets, often by a substantial margin. In contrast, PFN-based GFMs typically outperform GNNs: they achieve the best results on most of the datasets, with classic GNNs getting in top-$3$ only on $2$ out of the $10$ datasets (and even on these $2$ datasets, the first place belongs to GraphPFN).

    \vspace{-4pt}

    \item Comparing PFN-based GFMs and their different regimes and backbones to each other, we see that fine-tuning almost always improves the performance of these models (with the only exception being G2T-LimiX on \texttt{tolokers-2}), and these improvements are quite substantial. But even in the ICL regime, they always outperform all other GFMs, and also outperform all GNNs on the majority of datasets. The LimiX backbone is better than the TabPFNv2 backbone for TAG. In the ICL regime, when using the same LimiX backbone, G2T-FM and TAG (which both augment node features with graph-based information and then pass them directly to the tabular backbone) achieve close results (the same average rank, and G2T-FM slightly leads in the average normalized score). However, on $6$ out of the $10$ datasets, they are outperformed by GraphPFN (which directly incorporates graph message passing in the PFN architecture and thus natively supports graph-structured data) in the ICL regime. Further, in the fine-tuning regime, the lead of GraphPFN over G2T-FM significantly increases.

    \vspace{-4pt}

    \item GraphPFN in the fine-tuning regime is the best model overall, achieving the best results on all the $10$ datasets (often \hbox{outperforming all other models by a substantial margin}).


\end{itemize}


\vspace{-4pt}

\subsection{Results: Computational Efficiency}

\vspace{-4pt}

In this subsection, we discuss the computational costs of GNNs and PFN-based GFMs. First, it is important to note that these models operate in very different regimes, making a comparison of their computational costs not straightforward. GNNs follow the classic paradigm of requiring relatively long training (hundreds to thousands of forward and backward passes) to optimize model parameters for each specific dataset and then performing relatively fast inference (a single forward pass) to make predictions for this dataset. Further, as we have discussed, obtaining strong results from GNNs requires extensive hyperparameter tuning, i.e., performing many training runs (in our work, we perform $100$ training runs). The development and usage process of PFN-based GFMs is notably different. First, these models require a relatively expensive pretraining stage to obtain the base model. However, this pretraining is performed once by the model creators and then the same base model can be applied to different datasets, thus the pretraining costs are not relevant to the costs of model usage for a particular downstream application. After pretraining, PFNs can operate in the ICL regime without any dataset-specific training. However, as we have shown, dataset-specific fine-tuning of PFNs typically improves their performance. Thus, dataset-specific training is still often desirable. If dataset-specific training is performed, it requires selecting training hyperparameters, i.e., performing dataset-specific hyperparameter tuning. However, since the PFN architecture is completely fixed after pretraining, far fewer hyperparameters remain to be tuned than for GNNs trained from scratch (in our work, we perform $10$ training runs to select the learning rate).

For a representative subset of datasets, we provide the time required for the entire hyperparameter tuning, a single training run, and a single inference run of GNNs and PFN-based GFMs in Table~\ref{tab:time} (we provide a similar table for VRAM usage in Appendix~\ref{app:memory}). Note that if a PFN-based GFM is used in the ICL regime, then only the inference resources are relevant, since no dataset-specific training is performed.

\vspace{-2pt}

First, we emphasize that in industrial settings, which are arguably the most impactful real-world applications of node property prediction, inference time is what generally matters the most, since it is typically constrained by the specifics of the application such as the need for (almost) real-time prediction. We observe that in inference efficiency, GNNs strongly outcompete PFN-based GFMs: the inference time of GNNs is typically below $1$ second, while PFN-based GFMs require several seconds to several minutes, depending on the dataset (with inference time of TAG models sometimes reaching half an hour). The memory requirements of GNN inference are also typically significantly below those of PFN-based GFMs. However, when model development time rather than inference efficiency is critical, PFNs can be preferable to GNNs, as they can make predictions in the ICL regime without any dataset-specific fine-tuning, and even their fine-tuning can sometimes take less time than developing GNNs \hbox{since PFNs require less hyperparameter search}.

\vspace{-5pt}

\section{Conclusion}

\vspace{-5pt}

We performed a fair evaluation of a diverse set of GFMs on real-world node property prediction datasets and found that only PFN-based GFMs outcompete well-tuned classic GNNs, although at the cost of more expensive inference.

\bibliography{references}
\bibliographystyle{icml2026}


\newpage
\appendix
\onecolumn

\section{Non-PFN-Based GFMs for Node Property Prediction}
\label{app:gfms}

The key challenge for GFMs dealing with node property prediction is that graphs can come from vastly different domains (e.g., social networks, co-purchasing networks, road networks). This implies that the graphs can have different node feature spaces and node target spaces, i.e., have features and targets with different dimensions, distributions, meanings, and relations to each other. Thus, a successful GFM for node property prediction must be able to effectively adapt to graphs with different node feature and target spaces. The GFMs being proposed in the literature differ significantly in their mechanisms for this adaptation. In this section, we briefly describe a few common examples of these mechanisms. For tackling different node feature spaces, some GFMs construct additional synthetic edges based on node feature vector similarity and then discard node features \citep{OpenGraph}, while others only work with text-attributed graphs, i.e. graphs with textual descriptions as node features, which allows using Large Language Models (LLMs) to project these features to a common latent space \citep{OFA, li2024zerog, UniGraph}. However, such approaches cannot work effectively with all graphs, as many real-world graphs come with diverse numerical and categorical features that cannot be effectively processed by current LLMs and are too valuable for the task to discard. Other approaches use dimensionality reduction techniques like SVD or PCA to project all features into a space of the same dimensionality \citep{AnyGraph, GCOPE, MDGFM, SAMGPT}. However, the obtained spaces for different datasets are only matched in dimensionality, not in semantics. Another approach is to use a lightweight input module trained for each dataset from scratch \citep{GraphFM}, which can work well in the fine-tuning regime, but does not allow making predictions in the in-context learning regime. Mechanisms for adapting to different target spaces also vary. Some methods cast node classification as link prediction relative to virtual class nodes \citep{OpenGraph, AnyGraph}, however, this does not support node regression. A different approach uses a lightweight output module trained for each dataset from scratch \citep{GraphFM}, however, this does not support making predictions in the in-context learning regime.

\section{Prior-Data Fitted Networks (PFNs)}
\label{app:pfns}

The Prior-data Fitted Networks (PFNs) paradigm was introduced by \citet{muller2022transformers} (similar ideas concurrently appeared in \citep{nguyen2022transformer} under the name of Transformer Neural Processes). The idea of PFNs is to train models to make predictions on previously unseen datasets in a single forward pass. PFNs perform in-context learning (ICL): rather than updating the model parameters for each new dataset, they use the context provided as input to make predictions without per-dataset training. In the PFNs paradigm, the input to the model is an entire dataset that is split into two parts: a set of training samples paired with their labels (the \textit{context}), and a set of test samples without labels for which the predictions should be made (the \textit{query}). During the forward pass, the model uses the context samples to make predictions for the query samples, thus performing ICL. At the end of the forward pass, the model directly outputs predictions for the query samples. PFNs use (variants of) the Transformer architecture \citep{vaswani2017attention}. Each sample in the dataset is represented as one or more tokens. Samples interact through inter-sample attention that uses a specific attention pattern: context (training) samples are allowed to attend to all other context samples but not query samples, while query (test) samples are allowed to attend only to context samples and not to each other. This attention pattern means that predictions for each query sample are based only on this query sample and the context samples. A PFN is trained on a large number of diverse datasets to make its ICL mechanism work. At each training step, a PFN receives several new datasets, makes predictions for their query samples in a single forward pass, these predictions are used for loss computation, then the backward pass is performed to compute model parameter gradients, the parameters are updated based on these gradients concluding the training step, and the next training step uses a different set of datasets. To ensure a sufficient number and diversity of training datasets, these datasets are typically synthetically generated. A \textit{prior distribution} over datasets is defined, and synthetic datasets are sampled from this distribution. The design of the prior distribution can strongly affect the predictive performance of the PFN. \citet{muller2022transformers} showed that the PFN training procedure trains the model to approximate the posterior predictive distribution under the chosen prior distribution, which provides a theoretical motivation for the PFNs paradigm. PFNs can also be viewed as a meta-learning (i.e., ``learning how to learn'') approach that learns how to perform in-context learning. 

The PFNs paradigm was used to create successful Foundation Models for tabular data such as TabPFN \citep{TabPFN, TabPFNv2, TabPFNv3}, TabICL \citep{qu2025tabicl, qu2026tabiclv2}, and LimiX \citep{LimiX, LimiX-2M}. Inspired by this, several works have adopted this paradigm to create GFMs for node property prediction \citep{G2TFM, eremeev2026graphpfn, choi2025can, NodePFN, TAG}. The earlier of these models~--- G2T-FM \citep{G2TFM}, TabPFN-GN \citep{choi2025can}, and TAG \citep{TAG}~--- cast graph node classification/regression tasks as tabular classification/regression tasks by augmenting node features with aggregated graph neighborhood features and graph structure information, then discarding the graph structure, and directly applying unchanged tabular Foundation Models to the obtained feature tables. These three works mostly differ in the way they construct the additional graph-based features. Later PFN-based GFMs~--- GraphPFN \citep{eremeev2026graphpfn} and NodePFN \citep{NodePFN}~--- are significantly more involved: they create custom graph-native model architectures by augmenting standard PFN Transformers with graph neighborhood aggregation modules, design prior distributions over attributed graphs, and train these models on synthetic node property prediction datasets sampled from these prior distributions. Thus, these models handle graph-structured data natively instead of having to simplify it by converting to a table. GraphPFN and NodePFN differ in their model architecture and prior distribution designs, as well as in the fact that GraphPFN initializes non-graph-specific parts of the model from the Tabular Foundation Model LimiX-16M \citep{LimiX}, while NodePFN trains all parameters in the model from scratch. Note that in our work we do not evaluate TabPFN-GN, since there is no publicly available implementation of it, and NodePFN, since it has been shown by \citet{eremeev2026graphpfn} that NodePFN is strongly outperformed by GraphPFN (we hypothesize that this is due to some or all of the following: NodePFN uses an older architecture based on TabPFN while GraphPFN uses a newer architecture based on LimiX-16M which is in turn based on TabPFNv2, NodePFN is pretrained from scratch while GraphPFN initializes most of its parameters from LimiX-16M, and GraphPFN uses a significantly more complex prior distribution over graph datasets than NodePFN).

PFNs are designed specifically to be able to make predictions in the in-context learning regime. However, it was shown both for PFN-based Tabular Foundation Models \citep{rubachev2025finetuning} and Graph Foundation Models \citep{G2TFM, eremeev2026graphpfn} that their prediction quality can often be improved by additionally fine-tuning them on the downstream dataset.

Our experimental results in Section~\ref{sec:experiments} demonstrate that PFN-based GFMs are currently the best GFMs for node property prediction both in the in-context learning and fine-tuning regimes, and are the only current GFMs capable of outperforming well-tuned classic GNNs.

\section{Improved GNNs}
\label{app:gnns}

We believe the predictive performance of classic GNNs is often underestimated due to using weak GNN backbones. Many GNN implementations do not include such standard deep learning building blocks as skip-connections \citep{he2016deep} and normalizations \citep{ioffe2015batch, ba2016layer}, which often significantly improve the performance of neural models and are frequently used by the models being compared to GNNs. Further, from our extensive experience of training GNNs across a wide range of datasets, tasks, and settings, both for research and for industrial applications, we observe that obtaining optimal performance with GNNs requires extensive hyperparameter tuning, which is often not performed in prior works. Thus, in this work, to fairly represent the capabilities of GNNs, we use GNN implementations with improved backbones from \citet{platonov2023critical} and \citet{CGASB} and perform extensive per-dataset hyperparameter search for each of the considered GNNs. The GNN implementations from \citet{platonov2023critical} include skip-connections \cite{he2016deep}, 2-layer MLPs between graph message passing modules, and an option to use layer normalization \citep{ba2016layer}, or batch normalization \citep{ioffe2015batch}, or no normalization (this option is treated as a hyperparameter). The GNN implementations from \citet{CGASB} include an option to use layer normalization \citep{ba2016layer}, or batch normalization \citep{ioffe2015batch}, or no normalization, and an option to use ego- and neighbor-embedding separation (which they refer to as a linear residual connection), which allows the model to concatenate the embedding of the ego-node to the result of aggregating embeddings of its neighbors during graph message passing, thus preserving more information about the ego-node in its combined representation \citep{hamilton2017inductive, zhu2020beyond, platonov2023critical}. All the considered GNNs also include dropout \citep{srivastava2014dropout}. For all the considered GNNs, we perform $100$ trials of hyperparameter search with the popular TPE algorithm \citep{bergstra2011algorithms, watanabe2023tree}, see Appendix~\ref{app:experimental-setup} and Appendix~\ref{app:hparams} for more details.

More specifically, we consider $4$ GNN variants and have $2$ versions of each of these $4$ variants: one following the implementation from \citet{platonov2023critical} and one following the implementation from \citet{CGASB}. These $4$ variants are $2$ classic GNN models~--- GCN \citep{kipf2017semi} and GraphSAGE \citep{hamilton2017inductive}~--- which both use fixed graph neighborhood aggregation (message passing) operations, and $2$ GNN models with adaptive attention-based graph neighborhood aggregation operations~--- GAT \citep{velivckovic2018graph} with simple additive attention and Local Graph Transformer (LGT) \citep{shi2021masked, platonov2023critical} with scaled dot product attention \citep{vaswani2017attention} (note that this Graph Transformer variant only allows each node to attend to its neighbors~--- we specifically refer to it as a Local Graph Transformer following \citet{platonov2026cluster} to distinguish it from more common Global Graph Transformers that allow each node to attend to any other node unconstrained by the graph structure). The official implementation of GNNs from \citet{platonov2023critical} includes all $4$ of these GNN variants, while the official implementation of GNNs from \citet{CGASB} includes $3$ of them~--- it does not have an implementation of LGT, which we add by introducing an option to use the $\operatorname{TransformerConv}$ module from PyTorch Geometric \citep{fey2019fast} in the $\operatorname{MPNNs}$ class from the official codebase of \citet{CGASB}.

In our experiments in Section~\ref{sec:experiments}, we observe that the GNN implementations from \citet{CGASB} typically perform better for the simpler GCN and GraphSAGE models, while the GNN implementations from \citet{platonov2023critical} typically perform better for the more complex attention-based GAT and LGT models. The GAT implementation from \citet{platonov2023critical} (referred to as Crit.-GAT in Tables~\ref{tab:experimental-results}, \ref{tab:time}, \ref{tab:memory}) is the strongest of the GNNs in our experiments, achieving the fifth place among all the considered models according to both average rank and average normalized score aggregated metrics (and the third and the second places on the subset of only node classification datasets according to average rank and average normalized score aggregated metrics, respectively). Further, we also observe that the GNN implementations from \citet{platonov2023critical} tend to be more computationally efficient (often substantially) in both required time and memory usage than the GNN implementations from \citet{CGASB}. This is due to graph message passing in GNNs from \citet{platonov2023critical} being implemented using DGL \citep{wang2019deep}, while graph message passing in GNNs from \citet{CGASB} is implemented using PyTorch Geometric \citep{fey2019fast}: we find that DGL typically allows more efficient implementation of graph message passing than PyTorch Geometric (unfortunately, to the best of our knowledge, DGL is no longer being actively developed).

\section{Dataset Selection}
\label{app:datasets}

Reliable evaluation of GFMs for node property prediction requires careful dataset selection. Since the key characteristic of Foundation Models is that they can work well across diverse settings, the selected datasets should be diverse. In node property prediction, there are several potential dimensions of diversity, and it is desirable to cover all of them: diversity in data domains (e.g., social networks, web graphs, co-purchasing networks, road networks), diversity in tasks (e.g., fraud detection, view count prediction, topic prediction, CTR prediction), diversity in graph structure (size, sparsity, clustering, average shortest path length), diversity in graph structure and node labels relationships (homophily/assortativity~--- see \citet{newman2003mixing, platonov2023characterizing, mironov2024revisiting} for how to reliably measure this property), diversity in node features (features with different dimensions, distributions, and meanings).

Historically, standard benchmarks in graph node property prediction have been mostly limited to topic prediction in academic citation networks (perhaps due to the ease of obtaining such datasets), which obviously does not offer the necessary diversity~--- these datasets represent a single domain and task, have very similar graph structures, similarly high homophily, and typically come with outdated text-based features such as bag-of-words, TF-IDF, or bag-of-word-embeddings. Recently, standard Graph ML benchmarks have been extensively criticized \citep{bechler2025position, GraphLand}. Besides extensive reliance on citation networks and general low diversity, other notable points of critique include questionable practical relevance of many datasets, lack of complex and challenging tasks, limited representation of non-textual features common in real-world applications, unclear benefits of the graph structure for the considered tasks, potential bugs introduced in some datasets during data collection (we describe a number of previously reported and new problems with commonly used datasets in Appendix~\ref{app:dataset-problems}). We believe these issues call for a change in standard datasets used for benchmarking Graph ML methods, and the rise of interest in developing GFMs provides a convenient opportunity to realize such a change in benchmarking together with a change in the types of models being benchmarked.

Due to increasing concerns about the quality of standard Graph ML datasets, two new benchmarks aiming for high quality, diversity, and practical relevance have been recently proposed: GraphLand \citep{GraphLand} and GraphBench \citep{stoll2025graphbench}. While GraphBench mostly aims at widening the scope of Graph ML tasks and does not present many classic node property prediction tasks, GraphLand aims exactly at node property prediction and provides a diverse collection of datasets representing real-world industrial applications of this task. We believe GraphLand can be a much better testbed for evaluating GFMs for node property prediction than those commonly used in the current literature, and also encourage researchers and practitioners to release more datasets representing real-world applications of Graph ML. In our work, we use the $10$ sub-million-node datasets from GraphLand for benchmarking GFMs for node property prediction (we do not consider the $4$ million-node-scale datasets as none of the considered GFMs can work with datasets of such size, which highlights the importance of developing more scalable GFMs as a future direction of research).

\section{Experimental Setup}
\label{app:experimental-setup}

\vspace{-1pt}

For all datasets from the GraphLand benchmark, we use the official \texttt{RL} (random low) data splits, which are $10\%/10\%/80\%$ random stratified train/val/test splits. For each dataset and model, we report the mean and standard deviation of the obtained results computed over 10 runs. Graphs in some of the considered datasets are undirected and in some are directed; we preprocess the directed graphs by converting them to undirected, as done for the original GraphLand benchmark experiments \citep{GraphLand}.

\vspace{-1pt}

Hyperparameter selection is extremely important for achieving strong performance with GNNs, and very different hyperparameters can be optimal for different datasets. Thus, for each of the considered GNNs, we run $100$ trials of hyperparameter optimization for each dataset using the popular Tree-structured Parzen Estimator (TPE) algorithm \citep{bergstra2011algorithms, watanabe2023tree} provided in the Optuna library \citep{akiba2019optuna}. For each of the considered GNNs, we search over all hyperparameters available in the corresponding official implementation. The complete hyperparameter search spaces are provided in Appendix~\ref{app:hparams}. We train all GNNs with the AdamW optimizer \citep{kingma2014adam, loshchilov2019decoupled} for a maximum of $3000$ steps using early stopping based on the validation set performance with a patience of $1000$ steps.

\vspace{-1pt}

GFMs can support the in-context learning (ICL) regime, the fine-tuning (FT) regime, or both. One of the benefits of the ICL regime is that it does not require hyperparameter optimization, as the pretrained model is used as is without weight optimization. In the FT regime, hyperparameter search for GFMs can be useful. However, on the one hand, many hyperparameter search trials often cannot be afforded, as GFMs are typically much more computationally expensive than classic GNNs, and, on the other hand, the hyperparameter search space is naturally much smaller, as the architecture of the model is fixed, and the only significant tunable hyperparameter is typically the learning rate. Thus, for GFMs in the FT regime, we only tune the learning rate, for which we greedily search over a set of $10$ values. We follow the official implementations of the considered GFMs for other details of the fine-tuning strategy.

\vspace{-1pt}

Some GFMs incorporate randomness in their inference, which allows for inference-time ensembling, i.e., running inference multiple times and aggregating predictions. Among the GFMs evaluated by us, this technique is supported and used by default by PFN-based GFMs: G2T-FM, TAG, and GraphPFN. We provide more details on inference-time ensembling and its influence on prediction quality of PFN-based GFMs in Appendix~\ref{app:ensembling}.

\vspace{-1pt}

We use the official implementations provided by the authors for all GNNs and GFMs. All models used in our experiments are implemented with PyTorch \citep{paszke2019pytorch}, DGL \citep{wang2019deep}, and PyTorch Geometric \citep{fey2019fast}. All experiments were run on NVIDIA Tesla A100 80GB GPUs. Note that, due to the specifics of our infrastructure, our time measurements reported in Table~\ref{tab:time} may have noise up to $15\%$, but this does not affect our conclusions, as differences between time measurements for different model classes (GNNs and PFN-based GFMs) are significantly larger than that.

\vspace{-1pt}

\section{Memory Requirements of GNNs and PFN-Based GFMs}
\label{app:memory}

In Table~\ref{tab:memory}, we provide the VRAM required for training and inference of GNNs and PFN-based GFMs (note that, when numbers above $80$GB are reported, training on multiple GPUs with model parallelism was used). It can be seen that PFN-based GFMs are typically significantly more memory-intensive than GNNs, which is due to PFNs treating each feature as a separate token and thus creating a separate vector representation for each feature (while GNNs create a single vector representation for all features of a single node), and also due to PFNs often using larger models.

\vspace{-1pt}

\section{Inference-Time Ensembling of PFN-Based GFMs}
\label{app:ensembling}

\vspace{-2pt}

\begin{table}[t]
\centering
\caption{The memory (VRAM, in GB) required for: a single training run with the best hyperparameters (Tr), a single inference run with the best hyperparameters (Inf). Note that for GNNs, the models with the best hyperparameters can be widely different in size across GNN types and datasets. The \textit{Crit.} and \textit{Class.} prefixes refer to improved GNNs from \citet{platonov2023critical} and \citet{CGASB}, respectively (by the first words of the paper titles).}
\input{inputs/tables/total_memory}
\label{tab:memory}
\end{table}

\begin{table}[t]
\centering
\caption{The difference in predictive performance of PFN-based GFMs when making a single inference run and when using inference-time ensembling of $10$ inference runs.}
\input{inputs/tables/ensembling}
\label{tab:ensembling}
\vspace{-8pt}
\end{table}

Some models incorporate randomness in their inference, which allows for inference-time ensembling, i.e., running inference multiple times and aggregating predictions, which can lead to better prediction quality, although at a higher computational cost. This technique is commonly used in models following the PFNs paradigm, and is used by default by all the considered PFN-based GFMs: G2T-FM, TAG, and GraphPFN. We follow the official implementations of these models. For G2T-FM and GraphPFN, the official implementations use $10$ inference runs for ensembling. In this section, we additionally investigate how this inference-time ensembling influences the predictive performance of these models. We only consider G2T-FM and GraphPFN models here, as TAG uses a more complicated ensembling strategy and, besides using multiple runs of the Tabular Foundation Model backbone, also includes in its ensemble a number of `linear GNN' models (which are linear models fitted to predict node targets from node features augmented with various graph-neighborhood-aggregated features).

\vspace{-1pt}

We report the results obtained on a subset of datasets by a single inference run and $10$ inference runs of G2T-LimiX and GraphPFN in Table~\ref{tab:ensembling}. We can see that inference-time ensembling almost always improves the predictive performance of the model (there are exceptions in $2$ cases for G2T-LimiX), and these improvements range from very minor to quite substantial. However, even without inference-time ensembling, the performance of these models remains very strong. In the ICL regime, G2T-LimiX and GraphPFN outperform the best considered GNN on $3$ and $6$ out of the $8$ datasets, respectively, and in the FT regime, G2T-LimiX and GraphPFN outperform the best considered GNN on $7$ and $8$ out of the $8$ datasets, respectively. We note that, when run sequentially (like in our case) rather than in parallel, such inference-time ensembling increases inference time tenfold (although it does not affect memory requirements). Overall, inference-time ensembling provides a prediction quality/time trade-off, and whether using it is worth it depends on the particular application.

\vspace{-1pt}

\section{Problems of Commonly Used Node Property Prediction Datasets}
\label{app:dataset-problems}

\vspace{-1pt}

Historically, datasets for node property prediction in Graph ML have been dominated by academic citation networks, perhaps due to the ease of obtaining such datasets from open sources. Many works evaluate their methods predominantly or even exclusively on such datasets. However, this practice significantly limits the diversity of the evaluation setting: most citation network datasets are very similar to each other, and they also represent only a single data domain (which is also not particularly relevant to any real-world impactful applications). Further, the task most typically performed on these datasets is paper topic prediction; however, the topic of a paper can often be not uniquely defined, and it has been shown that a lot of nodes in these datasets appear to be mislabeled \citep{li2023graphcleaner}.

\vspace{-1pt}

There are also problems with other commonly used datasets for node property prediction. Some of them do not represent realistic applications (\texttt{coauthor-cs}, \texttt{coauthor-physics}, \texttt{airports-usa}, \texttt{airports-brazil}, \texttt{airports-europe}), in some, the provided graph does not appear to be beneficial for the considered task (\texttt{actor}), some come with no (or almost no) node features (\texttt{airports-usa}, \texttt{airports-brazil}, \texttt{airports-europe}, \texttt{deezer-hu}, \texttt{deezer-hr}, \texttt{deezer-ro}, \texttt{ogbn-proteins}), some are synthetic or semi-synthetic (\texttt{minesweeper}, \texttt{roman-empire}). Popular heterophilous node property prediction datasets have been criticized by \citet{platonov2023critical} who point out that some of these datasets are extremely small and provide insufficient class representation (\texttt{texas}, \texttt{cornell}, \texttt{wisconsin}, with the \texttt{texas} dataset having a class that consists of a single node), while others appear to have a large number of duplicated nodes, which indicates a bug in the data collection process and leads to train-test data leakage (\texttt{squirrel}, \texttt{chameleon}). While \citet{platonov2023critical} created versions of these datasets with duplicated nodes removed, the purpose of these versions was to empirically demonstrate that the presence of duplicated nodes leads to biased evaluation results, rather than to make these filtered versions new standard benchmarks, since the exact reason duplicated nodes appear is not known and thus it is also not known whether the datasets with these nodes removed are meaningful and useful; yet some works use these filtered dataset versions for evaluation.

\vspace{-1pt}

The problem with benchmarking in node property prediction is exacerbated by the fact that most datasets do not come with standardized data splits. As a consequence, different data splits are frequently used for the same datasets (often without being explicitly specified), thus complicating model comparison between different works, and some works borrow results reported elsewhere while using a different data split for their own experiments.

\vspace{-1pt}

\section{Hyperparameter Search Spaces for GNNs}
\label{app:hparams}

\vspace{-1pt}

In this section, we provide the complete hyperparameter search spaces for all the considered GNNs and GFMs. We use the following notation for distributions:

\vspace{-6pt}

\begin{itemize}[leftmargin=16pt]
    \item $\mathrm{Int}[\texttt{low}, \texttt{high}, \texttt{step}]$~--- integer hyperparameter sampled from a uniform distribution from \texttt{low} to \texttt{high} with step \texttt{step}.

    \vspace{-1pt}

    \item $\mathrm{Float}[\texttt{low}, \texttt{high}, \texttt{step}]$~--- real hyperparameter sampled from a uniform distribution from \texttt{low} to \texttt{high} with step \texttt{step}.

    \vspace{-1pt}

    \item $\mathrm{LogUniform}[\texttt{low}, \texttt{high}]$~--- real hyperparameter sampled from a log-uniform distribution from \texttt{low} to \texttt{high}.

    \vspace{-1pt}

    \item $\mathrm{Cat[\texttt{a},\ \texttt{b},\ \texttt{c},\ \dots]}$~--- categorical hyperparameter sampled from a uniform distribution on unordered values \texttt{a}, \texttt{b}, \texttt{c}, \dots.
\end{itemize}

\vspace{-6pt}

The complete hyperparameter search spaces are provided in Table~\ref{tab:hparams-crit} for GNN implementations from \citet{platonov2023critical} and in Table~\ref{tab:hparams-class} for GNN implementations from \citet{CGASB}.

\vspace{-1pt}

We note that \texttt{identity\_aggregation} refers to concatenating the pre-aggregation node embedding to the output of the neighborhood aggregation operation, since it essentially uses the identity function to propagate the node representation further down the model (this technique popularized by \citet{hamilton2017inductive} is also sometimes referred to as ego- and neighbor-embedding separation \citep{zhu2020beyond, platonov2023critical}). The other hyperparameter names are self-explanatory.

\vspace{-1pt}

\begin{table}[!ht]
\centering
\caption{The hyperparameter search distributions for Crit.-GNNs, i.e., GNN implementations from \citet{platonov2023critical}.}
\input{inputs/tables/hparams_crit}
\label{tab:hparams-crit}
\vspace{-3pt}
\end{table}

\begin{table}[!ht]
\centering
\caption{The hyperparameter search distributions for Class.-GNNs, i.e., GNN implementations from \citet{CGASB}.}
\input{inputs/tables/hparams_class}
\label{tab:hparams-class}
\end{table}

We used these hyperparameter search spaces for all GNN models and on all datasets, with one minor deviation which we describe here. We found that LGT models can sometimes be very unstable with some normalization choices, and this can occasionally lead to the hyperparameter optimization procedure selecting highly suboptimal hyperparameters. Thus, we reduced the set of the available normalization options for LGT models on several datasets. Specifically, for the LGT implementation from \citet{platonov2023critical} (Crit.-LGT) we removed the option to use no normalization on the \texttt{artnet-views} dataset, and for the LGT implementation from \citet{CGASB} (Class.-LGT) we removed the option to use BatchNorm on the \texttt{hm-categories} and \texttt{twitch-views} datasets.

\section{Limitations}

Our study reevaluates 9 GFMs for node property prediction from prior literature, covering several distinct model families. However, GFM development is an active research area, and new models continue to appear regularly. Thus, our evaluation is not exhaustive and, in particular, may omit recently proposed or concurrently developed methods.

Our experiments are conducted on datasets from the recently introduced GraphLand benchmark~\citep{GraphLand}. In Appendix~\ref{app:datasets}, we motivate this choice and discuss limitations of other existing node property prediction datasets. We believe GraphLand is currently the most suitable benchmark for node property prediction evaluation because it covers diverse and practical node property prediction tasks. Nevertheless, our conclusions may not generalize to domains outside GraphLand, such as proprietary datasets, highly specialized application areas, or datasets with different feature types or distributions. 

Despite these limitations, to the best of our knowledge, our work provides the most comprehensive evaluations of GFMs for node property prediction to date.

\section{Reproducibility}

The code for reproducing our experiments with improved GNN implementations from \citet{platonov2023critical} and \citet{CGASB} is available in \href{https://github.com/yandex-research/gnn-fair-evaluation/}{this GitHub repository}. For each of the considered GFMs, the official codebase for the respective model was used (with support for datasets from the GraphLand benchmark \citep{GraphLand} added by us where necessary).


\end{document}

%% file: inputs/tables/results.tex
\resizebox{\textwidth}{!}{
\begin{tabular}{lcccccccccccccc}
& \multicolumn{1}{c}{mult. class.} & \multicolumn{3}{c}{bin. class.} & \multicolumn{6}{c}{regression} \\
\cmidrule(lr){2-2}
\cmidrule(lr){3-5}
\cmidrule(lr){6-11}
& \texttt{\footnotesize hm-categories} & \texttt{\footnotesize tolokers-2} & \texttt{\footnotesize city-reviews} & \texttt{\footnotesize artnet-exp} & \texttt{\footnotesize hm-prices} & \texttt{\footnotesize avazu-ctr} & \texttt{\footnotesize city-roads-M} & \texttt{\footnotesize city-roads-L} & \texttt{\footnotesize twitch-views} & \texttt{\footnotesize artnet-views} & rank (cls) $\downarrow$ & rank (all)  $\downarrow$ & score (cls) $\uparrow$ & score (all) $\uparrow$ \\
\midrule
Crit.-GraphSAGE & $64.03 \pm 1.87$ & $53.80 \pm 0.62$ & $77.54 \pm 0.15$ & $48.13 \pm 0.58$ & $75.02 \pm 0.09$ & $32.11 \pm 0.49$ & $59.69 \pm 0.54$ & $53.31 \pm 0.41$ & $71.37 \pm 0.53$ & $54.44 \pm 0.20$ & $11.00$ & $9.00$ & $83.17$ & $24.48$ \\
Class.-GraphSAGE & $69.09 \pm 0.52$ & $54.82 \pm 0.46$ & $78.40 \pm 0.14$ & $46.42 \pm 0.38$ & $73.39 \pm 0.32$ & $31.53 \pm 0.19$ & $60.15 \pm 0.58$ & $54.73 \pm 0.25$ & $73.52 \pm 0.12$ & $55.05 \pm 0.08$ & $10.00$ & $7.90$ & $85.09$ & $29.47$ \\
Crit.-GCN & $69.23 \pm 0.12$ & $53.52 \pm 0.61$ & $76.59 \pm 0.35$ & $45.46 \pm 0.74$ & $71.25 \pm 0.57$ & $31.48 \pm 0.16$ & $59.41 \pm 0.40$ & $52.06 \pm 0.78$ & \textcolor{violet}{$77.60 \pm 0.06$} & $57.53 \pm 0.31$ & $12.00$ & $9.20$ & $82.68$ & $23.82$ \\
Class.-GCN & $70.00 \pm 0.66$ & $54.68 \pm 0.40$ & $78.53 \pm 0.08$ & $48.15 \pm 0.29$ & $71.68 \pm 0.50$ & $30.98 \pm 0.33$ & $60.14 \pm 0.16$ & $55.34 \pm 0.20$ & \textcolor{orange}{$78.02 \pm 0.13$} & $58.87 \pm 0.32$ & $8.00$ & $6.60$ & $86.46$ & $37.73$ \\
Crit.-GAT & \textcolor{orange}{$73.46 \pm 0.27$} & $58.94 \pm 0.91$ & $78.42 \pm 0.18$ & $49.43 \pm 0.19$ & $75.43 \pm 0.22$ & $32.43 \pm 0.36$ & $59.68 \pm 0.42$ & $53.75 \pm 0.49$ & $75.26 \pm 0.30$ & $55.00 \pm 0.26$ & \textcolor{violet}{$5.25$} & $5.90$ & \textcolor{orange}{$91.54$} & $42.97$ \\
Class.-GAT & $70.05 \pm 0.89$ & $57.62 \pm 0.85$ & $78.40 \pm 0.10$ & $48.25 \pm 0.39$ & $67.17 \pm 0.10$ & $30.53 \pm 0.31$ & $60.22 \pm 0.47$ & $55.19 \pm 0.40$ & $73.63 \pm 1.22$ & $56.40 \pm 0.26$ & $7.00$ & $7.20$ & $88.61$ & $30.47$ \\
Crit.-LGT & \textcolor{violet}{$72.15 \pm 0.71$} & $56.20 \pm 0.32$ & $78.03 \pm 0.13$ & $49.15 \pm 0.67$ & $73.89 \pm 0.47$ & $31.44 \pm 0.57$ & $59.55 \pm 2.02$ & $54.12 \pm 0.29$ & $76.22 \pm 0.08$ & $53.89 \pm 0.21$ & $7.50$ & $7.70$ & $88.71$ & $35.16$ \\
Class.-LGT & $62.54 \pm 0.36$ & $56.94 \pm 0.27$ & $78.38 \pm 0.27$ & $48.04 \pm 0.45$ & $67.95 \pm 0.28$ & $30.09 \pm 0.20$ & $59.25 \pm 0.26$ & $54.71 \pm 0.27$ & $67.95 \pm 0.18$ & $54.24 \pm 1.24$ & $10.00$ & $10.20$ & $85.28$ & $17.22$ \\
\midrule
OpenGraph (ICL) & $9.88 \pm 0.76$ & $40.43 \pm 1.15$ & $58.54 \pm 1.60$ & $15.41 \pm 0.71$ & N/A & N/A & N/A & N/A & N/A & N/A & $17.25$ & N/A & $24.00$ & N/A \\
AnyGraph (ICL) & $14.15 \pm 1.64$ & $29.92 \pm 2.78$ & $64.54 \pm 1.67$ & $13.50 \pm 0.98$ & N/A & N/A & N/A & N/A & N/A & N/A & $18.25$ & N/A & $19.73$ & N/A \\
TS-GNN (ICL) & $20.09 \pm 1.29$ & $38.54 \pm 0.94$ & $36.85 \pm 15.65$ & $20.44 \pm 1.05$ & N/A & N/A & N/A & N/A & N/A & N/A & $16.25$ & N/A & $18.52$ & N/A \\
GCOPE (FT) & $19.57 \pm 0.12$ & $28.35 \pm 1.24$ & $66.11 \pm 0.74$ & $15.41 \pm 3.07$ & N/A & N/A & N/A & N/A & N/A & N/A & $17.50$ & N/A & $22.52$ & N/A \\
SAMGPT (FT) & $14.64 \pm 0.35$ & $36.87 \pm 1.04$ & $46.36 \pm 2.49$ & $16.16 \pm 0.87$ & N/A & N/A & N/A & N/A & N/A & N/A & $17.00$ & N/A & $17.46$ & N/A \\
MDGFM (FT) & $9.08 \pm 1.00$ & $31.24 \pm 1.71$ & $31.24 \pm 8.50$ & $15.52 \pm 1.46$ & N/A & N/A & N/A & N/A & N/A & N/A & $18.75$ & N/A & $3.36$ & N/A \\
\midrule
G2T-LimiX (ICL) & $58.05 \pm 0.28$ & \textcolor{orange}{$61.60 \pm 0.18$} & $78.98 \pm 0.44$ & $48.42 \pm 0.78$ & $76.14 \pm 0.08$ & \textcolor{violet}{$32.70 \pm 0.14$} & \textcolor{violet}{$65.16 \pm 0.07$} & $56.62 \pm 0.13$ & $71.31 \pm 0.06$ & $61.58 \pm 0.08$ & $6.25$ & $5.20$ & $87.59$ & $51.71$ \\
TAG-TabPFNv2 (ICL) & $57.17 \pm 0.46$ & $59.33 \pm 1.00$ & $77.38 \pm 0.29$ & $47.87 \pm 0.39$ & N/A & N/A & N/A & N/A & N/A & N/A & $11.00$ & N/A & $84.47$ & N/A \\
TAG-LimiX (ICL) & $58.69 \pm 0.39$ & $59.46 \pm 1.05$ & $78.87 \pm 0.13$ & $50.19 \pm 0.46$ & N/A & N/A & N/A & N/A & N/A & N/A & $6.25$ & N/A & $87.31$ & N/A \\
GraphPFN (ICL) & $55.25 \pm 0.75$ & \textcolor{violet}{$61.29 \pm 0.12$} & \textcolor{violet}{$80.25 \pm 0.05$} & \textcolor{orange}{$51.79 \pm 0.11$} & \textcolor{orange}{$77.88 \pm 0.09$} & $31.63 \pm 0.06$ & $64.85 \pm 0.13$ & \textcolor{violet}{$58.52 \pm 0.12$} & $73.20 \pm 0.08$ & \textcolor{violet}{$62.79 \pm 0.08$} & $5.50$ & \textcolor{violet}{$4.70$} & $89.10$ & \textcolor{violet}{$60.39$} \\
\midrule
G2T-LimiX (FT) & $65.59 \pm 0.21$ & $59.75 \pm 1.14$ & \textcolor{orange}{$80.65 \pm 0.05$} & \textcolor{violet}{$50.39 \pm 0.19$} & \textcolor{violet}{$77.37 \pm 0.17$} & \textcolor{orange}{$34.09 \pm 0.37$} & \textcolor{orange}{$66.29 \pm 0.12$} & \textcolor{orange}{$62.21 \pm 0.19$} & $74.91 \pm 0.06$ & \textcolor{orange}{$63.24 \pm 0.07$} & \textcolor{orange}{$4.25$} & \textcolor{orange}{$3.40$} & \textcolor{violet}{$91.02$} & \textcolor{orange}{$73.69$} \\
GraphPFN (FT) & \textcolor{red}{$78.67 \pm 0.20$} & \textcolor{red}{$62.80 \pm 0.39$} & \textcolor{red}{$80.90 \pm 0.03$} & \textcolor{red}{$53.49 \pm 0.81$} & \textcolor{red}{$81.06 \pm 0.24$} & \textcolor{red}{$35.07 \pm 0.34$} & \textcolor{red}{$67.30 \pm 0.21$} & \textcolor{red}{$64.12 \pm 0.27$} & \textcolor{red}{$79.00 \pm 0.14$} & \textcolor{red}{$65.35 \pm 0.06$} & \textcolor{red}{$1.00$} & \textcolor{red}{$1.00$} & \textcolor{red}{$100.00$} & \textcolor{red}{$100.00$} \\
\bottomrule
\end{tabular}
}

%% file: inputs/tables/total_time.tex
\resizebox{\textwidth}{!}{
\begin{tabular}{lccc|ccc|ccc|ccc|ccc|ccc}
& \multicolumn{3}{c}{\texttt{tolokers-2}} & \multicolumn{3}{c}{\texttt{city-reviews}} & \multicolumn{3}{c}{\texttt{artnet-views}} & \multicolumn{3}{c}{\texttt{city-roads-M}} & \multicolumn{3}{c}{\texttt{city-roads-L}} & \multicolumn{3}{c}{\texttt{hm-prices}} \\
\cmidrule(lr){2-4}
\cmidrule(lr){5-7}
\cmidrule(lr){8-10}
\cmidrule(lr){11-13}
\cmidrule(lr){14-16}
\cmidrule(lr){17-19}
& Tun & Tr & Inf & Tun & Tr & Inf & Tun & Tr & Inf & Tun & Tr & Inf & Tun & Tr & Inf & Tun & Tr & Inf \\
\midrule
Crit.-GraphSAGE & $2.16$h & $1.63$m & $0.05$s & $14.02$h & $11.70$m & $0.23$s & $4.08$h & $2.93$m & $0.03$s & $2.22$h & $1.44$m & $0.03$s & $6.70$h & $3.66$m & $0.06$s & $1.00$d & $15.44$m & $0.09$s \\
Class.-GraphSAGE & $17.49$h & $7.23$m & $0.06$s & $1.81$d & $15.38$m & $0.25$s & $20.68$h & $11.94$m & $0.06$s & $1.68$h & $35.74$s & $0.02$s & $6.82$h & $3.29$m & $0.06$s & $7.57$d & $1.69$h & $2.11$s \\
Crit.-GCN & $2.70$h & $1.47$m & $0.04$s & $7.11$h & $3.44$m & $0.13$s & $1.93$h & $33.73$s & $0.02$s & $1.77$h & $35.47$s & $0.02$s & $4.07$h & $4.13$m & $0.03$s & $19.59$h & $7.21$m & $0.05$s \\
Class.-GCN & $1.07$d & $7.84$m & $0.07$s & $1.88$d & $23.24$m & $0.22$s & $11.20$h & $4.62$m & $0.02$s & $2.59$h & $1.37$m & $0.02$s & $11.98$h & $8.24$m & $0.10$s & $1.59$d & $53.72$m & $0.44$s \\
Crit.-GAT & $3.22$h & $2.23$m & $0.05$s & $19.76$h & $14.30$m & $0.24$s & $2.77$h & $1.69$m & $0.02$s & $2.70$h & $1.77$m & $0.02$s & $8.18$h & $5.56$m & $0.04$s & $2.42$d & $36.54$m & $0.18$s \\
Class.-GAT & $1.45$d & $15.62$m & $0.06$s & $2.50$d & $35.30$m & $0.33$s & $19.74$h & $12.68$m & $0.05$s & $1.70$h & $54.22$s & $0.02$s & $17.75$h & $11.31$m & $0.11$s & $2.47$h & $21.08$m & $1.87$s \\
Crit.-LGT & $3.24$h & $2.03$m & $0.04$s & $20.78$h & $16.96$m & $0.30$s & $5.46$h & $4.64$m & $0.04$s & $4.63$h & $2.87$m & $0.04$s & $5.46$h & $2.32$m & $0.02$s & $3.99$d & $1.05$h & $0.37$s \\
Class.-LGT & $1.79$d & $9.23$m & $0.06$s & $1.89$d & $1.57$h & $1.19$s & $1.16$d & $14.16$m & $0.10$s & $5.72$h & $1.46$m & $0.02$s & $18.12$h & $20.08$m & $0.13$s & $1.20$h & $23.18$m & $0.84$s \\
\midrule
G2T-LimiX & $24.74$m & $2.56$m & $8.93$s & $13.02$h & $1.22$h & $6.63$m & $2.40$h & $22.65$m & $1.26$m & $1.35$h & $8.82$m & $38.27$s & $19.10$h & $1.42$h & $7.92$m & $6.87$h & $35.28$m & $6.45$m \\
TAG-TabPFNv2 & N/A & N/A & $1.90$m & N/A & N/A & $32.77$m & N/A & N/A & N/A & N/A & N/A & N/A & N/A & N/A & N/A & N/A & N/A & N/A \\
TAG-LimiX & N/A & N/A & $1.84$m & N/A & N/A & $36.60$m & N/A & N/A & N/A & N/A & N/A & N/A & N/A & N/A & N/A & N/A & N/A & N/A \\
GraphPFN & $40.25$m & $3.78$m & $6.01$s & $1.08$d & $2.50$h & $1.52$m & $8.49$h & $1.09$h & $29.49$s & $3.16$h & $17.05$m & $15.72$s & $21.01$h & $2.16$h & $58.52$s & $1.32$d & $3.49$h & $2.47$m \\
\bottomrule
\end{tabular}
}

%% file: inputs/tables/total_memory.tex
\resizebox{\textwidth}{!}{
\begin{tabular}{lcc|cc|cc|cc|cc|cc}
& \multicolumn{2}{c}{\texttt{tolokers-2}} & \multicolumn{2}{c}{\texttt{city-reviews}} & \multicolumn{2}{c}{\texttt{artnet-views}} & \multicolumn{2}{c}{\texttt{city-roads-M}} & \multicolumn{2}{c}{\texttt{city-roads-L}} & \multicolumn{2}{c}{\texttt{hm-prices}} \\
\cmidrule(lr){2-3}
\cmidrule(lr){4-5}
\cmidrule(lr){6-7}
\cmidrule(lr){8-9}
\cmidrule(lr){10-11}
\cmidrule(lr){12-13}
& Tr & Inf & Tr & Inf & Tr & Inf & Tr & Inf & Tr & Inf & Tr & Inf \\
\midrule
Crit.-GraphSAGE & $1.74$ & $0.44$ & $26.23$ & $3.10$ & $3.16$ & $0.41$ & $4.16$ & $0.65$ & $10.91$ & $1.93$ & $5.55$ & $1.14$ \\
Class.-GraphSAGE & $0.34$ & $0.32$ & $9.34$ & $5.33$ & $1.32$ & $0.75$ & $0.43$ & $0.21$ & $3.11$ & $1.44$ & $28.98$ & $28.29$ \\
Crit.-GCN & $0.34$ & $0.08$ & $1.50$ & $0.45$ & $0.43$ & $0.10$ & $0.64$ & $0.12$ & $3.69$ & $0.65$ & $3.17$ & $1.19$ \\
Class.-GCN & $3.24$ & $1.92$ & $16.25$ & $9.29$ & $0.90$ & $0.47$ & $1.81$ & $1.06$ & $8.57$ & $4.06$ & $25.10$ & $23.16$ \\
Crit.-GAT & $0.80$ & $0.10$ & $18.00$ & $1.77$ & $1.31$ & $0.34$ & $3.16$ & $0.58$ & $6.98$ & $0.74$ & $34.31$ & $2.89$ \\
Class.-GAT & $4.94$ & $1.28$ & $41.97$ & $12.34$ & $7.68$ & $2.19$ & $2.38$ & $1.02$ & $17.45$ & $3.45$ & $79.32$ & $39.52$ \\
Crit.-LGT & $0.94$ & $0.16$ & $31.68$ & $2.93$ & $3.87$ & $0.33$ & $7.45$ & $0.62$ & $2.09$ & $0.44$ & $10.36$ & $2.89$ \\
Class.-LGT & $4.51$ & $1.75$ & $77.82$ & $17.54$ & $12.66$ & $2.53$ & $1.81$ & $0.62$ & $17.11$ & $2.22$ & $51.50$ & $34.04$ \\
\midrule
G2T-LimiX & $16.57$ & $4.54$ & $53.05$ & $30.97$ & $25.55$ & $25.32$ & $21.04$ & $20.81$ & $73.17$ & $68.19$ & $64.20$ & $63.97$ \\
TAG-TabPFNv2 & N/A & $10.15$ & N/A & $29.17$ & N/A & N/A & N/A & N/A & N/A & N/A & N/A & N/A \\
TAG-LimiX & N/A & $17.75$ & N/A & $28.54$ & N/A & N/A & N/A & N/A & N/A & N/A & N/A & N/A \\
GraphPFN & $8.12$ & $1.63$ & $146.66$ & $31.95$ & $73.60$ & $14.97$ & $41.78$ & $8.30$ & $145.49$ & $36.34$ & $70.81$ & $20.52$ \\
\bottomrule
\end{tabular}
}

%% file: inputs/tables/ensembling.tex
\resizebox{\textwidth}{!}{
\begin{tabular}{lcccccccc}
\toprule
& \texttt{\footnotesize tolokers-2} & \texttt{\footnotesize city-reviews} & \texttt{\footnotesize artnet-exp} & \texttt{\footnotesize hm-prices} & \texttt{\footnotesize avazu-ctr} & \texttt{\footnotesize city-roads-M} & \texttt{\footnotesize twitch-views} & \texttt{\footnotesize artnet-views} \\
\midrule
Best GNN & $58.94 \pm 0.91$ & $78.53 \pm 0.08$ & $49.43 \pm 0.19$ & $75.43 \pm 0.22$ & $32.43 \pm 0.36$ & $60.22 \pm 0.47$ & $78.02 \pm 0.13$ & $58.87 \pm 0.32$ \\
\midrule
G2T-LimiX (ICL, 1 run) & $61.13 \pm 0.21$ & $77.29 \pm 0.54$ & $48.44 \pm 0.23$ & $75.41 \pm 0.04$ & $32.41 \pm 0.12$ & $64.53 \pm 0.09$ & $71.08 \pm 0.07$ & $60.95 \pm 0.09$ \\
G2T-LimiX (ICL, 10 runs) & $61.60 \pm 0.18$ & $78.98 \pm 0.44$ & $48.42 \pm 0.78$ & $76.14 \pm 0.08$ & $32.70 \pm 0.14$ & $65.16 \pm 0.07$ & $71.31 \pm 0.06$ & $61.58 \pm 0.08$ \\
\midrule
GraphPFN (ICL, 1 run) & $60.97 \pm 0.13$ & $80.01 \pm 0.03$ & $51.46 \pm 0.11$ & $77.33 \pm 0.06$ & $31.43 \pm 0.06$ & $64.03 \pm 0.11$ & $72.75 \pm 0.04$ & $62.07 \pm 0.05$ \\
GraphPFN (ICL, 10 runs) & $61.29 \pm 0.12$ & $80.25 \pm 0.05$ & $51.79 \pm 0.11$ & $77.88 \pm 0.09$ & $31.63 \pm 0.06$ & $64.85 \pm 0.13$ & $73.20 \pm 0.08$ & $62.79 \pm 0.08$ \\
\midrule
G2T-LimiX (FT, 1 run) & $60.82 \pm 0.57$ & $80.13 \pm 0.13$ & $49.79 \pm 0.21$ & $76.72 \pm 0.20$ & $34.03 \pm 0.35$ & $65.87 \pm 0.11$ & $74.31 \pm 0.13$ & $62.08 \pm 0.12$ \\
G2T-LimiX (FT, 10 runs) & $59.75 \pm 1.14$ & $80.65 \pm 0.05$ & $50.39 \pm 0.19$ & $77.37 \pm 0.17$ & $34.09 \pm 0.37$ & $66.29 \pm 0.12$ & $74.91 \pm 0.06$ & $63.24 \pm 0.07$ \\
\midrule
GraphPFN (FT, 1 run) & $62.72 \pm 0.24$ & $80.34 \pm 0.17$ & $53.33 \pm 0.23$ & $80.16 \pm 0.17$ & $34.89 \pm 0.16$ & $66.02 \pm 0.10$ & $78.46 \pm 0.09$ & $64.37 \pm 0.04$ \\
GraphPFN (FT, 10 runs) & $62.80 \pm 0.39$ & $80.90 \pm 0.03$ & $53.49 \pm 0.81$ & $81.06 \pm 0.24$ & $35.07 \pm 0.34$ & $67.30 \pm 0.21$ & $79.00 \pm 0.14$ & $65.35 \pm 0.06$ \\
\bottomrule
\end{tabular}
}

%% file: inputs/tables/hparams_crit.tex
\resizebox{0.7\textwidth}{!}{
\begin{tabular}{lll}
\toprule
Parameter & Distribution & Comment \\
\midrule
\texttt{learning\_rate} & $\mathrm{LogUniform}[0.00003, 0.01]$ \\
\texttt{weight\_decay} & $\mathrm{LogUniform}[0.0001, 1]$ \\
\texttt{dropout} & $\mathrm{Float}[0, 0.5, 0.05]$ \\
\texttt{num\_backbone\_blocks} & $\mathrm{Int}[1, 10, 1]$ \\
\texttt{hidden\_dim} & $\mathrm{Int}[96, 768, 32]$ \\
\texttt{normalization\_type} & $\mathrm{Cat}[\texttt{None},\ \texttt{LayerNorm},\ \texttt{BatchNorm}]$ \\
$\log_2(\texttt{num\_heads})$ & $\mathrm{Int}[1, 3, 1]$ & only for GAT and LGT models \\
\bottomrule
\end{tabular}
}

%% file: inputs/tables/hparams_class.tex
\resizebox{0.7\textwidth}{!}{
\begin{tabular}{lll}
\toprule
Parameter & Distribution & Comment \\
\midrule
\texttt{learning\_rate} & $\mathrm{LogUniform}[0.00003, 0.01]$ \\
\texttt{weight\_decay} & $\mathrm{LogUniform}[0.0001, 1]$ \\
\texttt{dropout} & $\mathrm{Float}[0, 0.5, 0.05]$ \\
\texttt{num\_backbone\_blocks} & $\mathrm{Int}[1, 10, 1]$ \\
\texttt{hidden\_dim} & $\mathrm{Int}[96, 768, 32]$ \\
\texttt{identity\_aggregation} & $\mathrm{Cat}[\texttt{False},\ \texttt{True}]$ \\
\texttt{normalization\_type} & $\mathrm{Cat}[\texttt{None},\ \texttt{LayerNorm},\ \texttt{BatchNorm}]$ \\
$\log_2(\texttt{num\_heads})$ & $\mathrm{Int}[1, 3, 1]$ & only for GAT and LGT models \\
\texttt{num\_input\_layers} & $\mathrm{Int}[0, 1, 1]$ \\
\bottomrule
\end{tabular}
}